\documentclass[runningheads]{llncs}
\usepackage{graphicx}
\usepackage[export]{adjustbox}

\usepackage[style=numeric]{biblatex}
\usepackage{xcolor}

\usepackage[colorlinks=true, linkcolor=black, citecolor=black, urlcolor=black]{hyperref}

\usepackage{booktabs} 
\usepackage{adjustbox} 
\usepackage{multirow} 

\usepackage{float}
\usepackage{subcaption}
\usepackage{amsfonts}
\usepackage{amsmath}
\usepackage{lineno}
\usepackage{afterpage}
\usepackage{orcidlink}
\usepackage{hyperref}

\begin{document}

\title{Leveraging Semantic Segmentation Masks with Embeddings for Fine-Grained Form Classification}

\author{Taylor Archibald\inst{1}\orcidlink{0000-0003-2576-208X} \and
Tony Martinez\inst{1}}

\titlerunning{Semantic Segmentation Masks for Fine-Grained Form Classification}
\authorrunning{Archibald and Martinez}

\institute{Brigham Young University, Provo, UT\\
\email{tarch@byu.edu, martinez@cs.byu.edu}\\
\url{https://axon.cs.byu.edu}
}
\maketitle
\begin{abstract}

Efficient categorization of historical documents is crucial for fields such as genealogy, legal research, and historical scholarship, where manual classification is impractical for large collections due to its labor-intensive and error-prone nature. To address this, we propose a representational learning strategy that integrates semantic segmentation and deep learning models—ResNets, CLIP, the Document Image Transformer (DiT), and masked auto-encoders (MAE)—to generate embeddings that capture document features without predefined labels. To the best of our knowledge, we are the first to evaluate embeddings on fine-grained, unsupervised form classification. To improve these embeddings, we propose to first employ semantic segmentation as a preprocessing step. We contribute two novel datasets—French 19th-century and U.S. 1950 Census records—to demonstrate our approach. Our results show the effectiveness of these various embedding techniques in distinguishing similar document types and indicate that applying semantic segmentation can greatly improve clustering and classification results. The census datasets are available at \href{https://github.com/tahlor/census_forms}{https://github.com/tahlor/census\_forms}.

\end{abstract}


\section{Introduction}
In classifying documents, the ability to categorize forms efficiently is essential. Forms often house critical information such as names, dates, and events, which are invaluable for various academic and professional fields, including genealogy, legal research, and historical scholarship. Classification enhances the utility of this data, enabling researchers to retrieve information more easily. Moreover, well-classified archives support more sophisticated digital searches, automated cross-referencing, and data integration tasks. Additionally, when historical records are well-organized, it becomes possible to estimate the volume of each document type, and the number of historical records contained on each document, which can further streamline their management. If these documents are being digitized, accurate classification enables automated systems to route documents to the appropriate processing workflows, optimizing both the efficiency and accuracy of data handling.

However, manual classification of historical records is time-consuming and error-prone due to the large volume and minor visual differences between forms. With collections often in the millions of documents, manual sorting is inefficient and costly. Automated systems are essential for efficient initial sorting and categorizing, reducing human workload and improving speed and accuracy in processing historical data.

To address these challenges, we propose a representational learning strategy that identifies various document types based on their visual attributes. We investigate the potential of deep learning approaches, including  ResNets (Residual Networks), CLIP (Contrastive Language-Image Pre-training), the Document Image Transformer (DiT), and masked auto-encoders (MAE)~\cite{heDeepResidualLearning2015,radfordLearningTransferableVisual2021,liDiTSelfsupervisedPretraining2022,heMaskedAutoencodersAre2021}. These models are adept at producing embeddings—dense vector representations—that capture the intrinsic features of each document type, mitigating the need for predefined category labels.

These embeddings can be utilized in unsupervised learning algorithms or visualized in 3D space. This visualization facilitates rapid, preliminary classification and supports data scientists in the evaluation process. By expediting the classification task, this approach not only enhances the speed of document processing but also ensures documents are directed to the most suitable processing pipelines for further analysis and handling.

We focus our investigation on representations capable of distinguishing between highly similar historical document types. We introduce two novel datasets derived from French 19th-century and U.S. 1950 census records, each characterized by subtly different form types. We assess the effectiveness of various embedding techniques in this nuanced context. Moreover, we demonstrate that removing handwritten text via semantic segmentation enhances these representations. The methodologies and insights from this study are poised to establish a new benchmark in unsupervised, fine-grained historical document classification.

\section{Related Work}
Representational learning has become a cornerstone in the advancement of machine learning and computer vision, particularly in tasks involving complex image data. This approach involves training models to learn compact, informative representations of data that capture essential features while discarding irrelevant information. Notably, Convolutional Neural Networks (CNNs) like ResNet have demonstrated significant success in image recognition by leveraging hierarchical feature extraction, which is further enhanced by pretraining on extensive datasets like ImageNet~\cite{donahueDeCAFDeepConvolutional2014, oquabLearningTransferringMidlevel2014, heDeepResidualLearning2015, dengImageNetLargescaleHierarchical2009}. The introduction of Vision Transformers (ViTs), particularly when trained as Masked Autoencoders (MAEs), marks a shift towards utilizing attention mechanisms to process image patches and capture global dependencies~\cite{dosovitskiyIMAGEWORTH16X16,heMaskedAutoencodersAre2021}. Extending this paradigm, the Document Image Transformer (DiT) targets document images, aiming to encapsulate both textual and visual elements effectively~\cite{liDiTSelfsupervisedPretraining2022}. Moreover, CLIP by OpenAI exemplifies the power of contrastive learning by aligning text and image embeddings in a shared latent space, thus enhancing performance across various image-based tasks~\cite{radfordLearningTransferableVisual2021}. These advancements collectively underscore the potential of deep learning models in learning robust and transferable representations, which are critical for fine-grained document classification.

While considerable work has been done in the domain of document classification, the goal has often been to classify images into predefined semantic document types rather than precise, but potentially unknown, form types~\cite{saifullahAnalyzingPotentialActive2022a,afzalCuttingErrorHalf2017,liDiTSelfsupervisedPretraining2022,harleyEvaluationDeepConvolutional2015,siddiquiSelfSupervisedRepresentationLearning2021,omurcaDocumentImageClassification2023}. For example, datasets like Tobacco-3482 and RVL-CDIP are commonly used for document classification, but they focus on broader classes of documents such as memos, emails, and advertisements~\cite{lewisBuildingTestCollection2006, harleyEvaluationDeepConvolutional2015}. In contrast, we are more interested in the fine-grained structural similarity of document images, distinguishing between forms, which may have identical content but slight variations in presentation.

\subsection{Semantic Segmentation}
A potential difficulty is that deep learning models can struggle to discern what is semantically important in varying contexts. However, preprocessing can enhance the model's ability to capture meaningful representations. We believe that semantic segmentation can play a crucial role in improving document representations. Semantic segmentation involves separating different components of the document, such as handwriting, printed text, and preprinted form elements, which can highlight the most relevant parts of the document for the model.

A classical form of semantic segmentation in documents is binarization, where pixels are classified into foreground or background to create a binary image representation. This problem, addressed in numerous studies over the years, saw significant advances through the Document Image Binarization Contest (DIBCO), held annually from 2009 to 2019. DIBCO has been pivotal in advancing document binarization techniques, providing a benchmarking platform for methods on diverse, challenging document images, including degraded historical manuscripts~\cite{pratikakisICDAR2017CompetitionDocument2017}. Deep learning methods, especially CNNs, have proven effective, surpassing traditional methods in adaptability to script variations and noise management~\cite{tensmeyerDocumentImageBinarization2017, ronnebergerUNetConvolutionalNetworks2015}. Notably, the U-Net architecture, adapted from medical imaging, showcased remarkable success in the 2017 DIBCO~\cite{ronnebergerUNetConvolutionalNetworks2015}.

\begin{figure}[h]
  \centering
  {\includegraphics[width=0.7\textwidth]{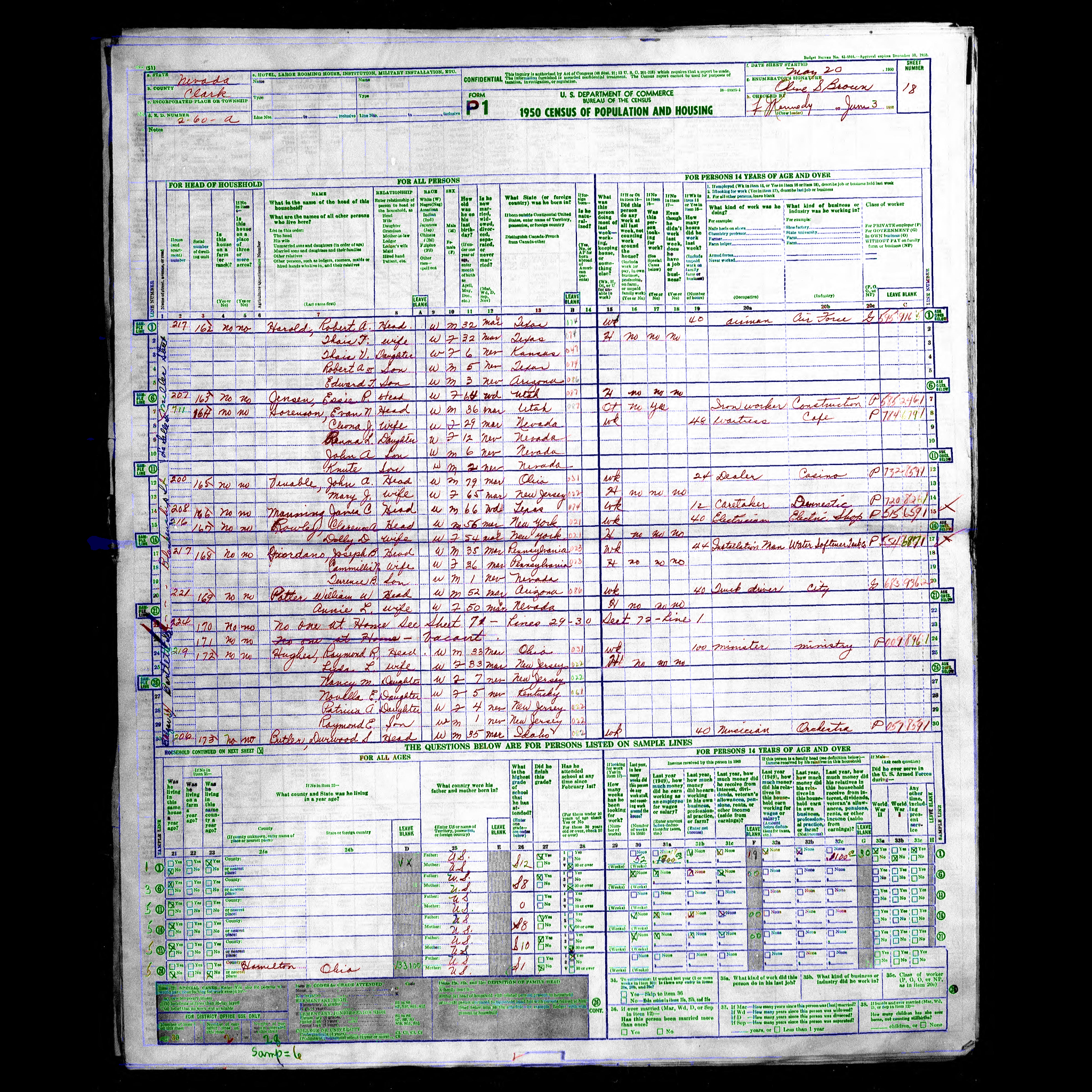}}
  \caption{An image of a U.S. 1950 Census form can be decomposed into different content classes using a model trained on DELINE8K. The handwriting has been tinted by class (handwriting=red, preprinted text=green, and grid lines=blue).
}
  \label{fig:1950_deline8k_segmentation}

\end{figure}

Multiclass semantic segmentation extends binarization to classify multiple document components, such as text, images, and form elements. However, this expansion faces challenges, primarily due to the labor-intensive nature of document labeling and the scarcity of comprehensive training datasets~\cite{stewartDocumentImagePage2017}. Researchers have addressed these challenges by creating synthetic datasets like the WGM-SYN, which blends archival printed texts with other document elements~\cite{vafaieHandwrittenPrintedText2022}, SignaTR6K, which combines legal document crops with superimposed signatures~\cite{gholamianHandwrittenPrintedText2023}, and DELINE8K, which offers a diverse array of classes including form elements, preprinted text, and handwriting, set against backgrounds generated by DALL-E that mimic historical documents~\cite{archibaldDELINE8KSyntheticData2024, rameshZeroShotTexttoImageGeneration}. Figure~\ref{fig:1950_deline8k_segmentation} shows a U.S. 1950 Census form with the semantic segmentation visualized generated from a model trained on DELINE8K.

These semantic segmentation approaches can be integrated with other representational learning strategies to enhance the model's ability to focus on the most pertinent features. By employing semantic segmentation as a preprocessing step, we isolate essential components of each document, such as handwritten text, printed text, and preprinted form elements, enabling the model to concentrate on the most informative aspects.

Our approach leverages this preprocessing to improve unsupervised form classification. While traditional methods classify documents into broad categories, fine-grained classification of forms with slight presentation variations remains challenging, especially in unsupervised scenarios. To our knowledge, no prior work has combined semantic segmentation with unsupervised representational learning for detailed form classification.

This novel integration aims to set a new benchmark, demonstrating that semantic segmentation can significantly enhance unsupervised learning models in capturing and utilizing the structural nuances of historical documents.

\section{Methods}
The primary goal of our research is to compute robust representations for classifying historical document images based on their structural characteristics. Mathematically, we aim to define a function \( f: \mathcal{X} \rightarrow \mathcal{Y} \) that maps an input image \( x \in \mathcal{X} \) to a feature space \( \mathcal{Y} \) that captures essential visual features for classification. The quality of the learned representations is evaluated based on their ability to cluster documents into their correct categories without prior knowledge of these categories.

\subsection{Evaluation}

In evaluating our representational learning models, we focus on the classification loss to ensure that the representations are not only separating but also informative for classification tasks. For models tested with a linear classifier (linear probe), we minimize the cross-entropy loss.
Following~\cite{heMaskedAutoencodersAre2021}, we pass the embedding into a batch-normalized fully connected layer. We train it using the AdamW~\cite{loshchilovDecoupledWeightDecay2018} optimizer with a step decay learning rate schedule. We perform 10-fold cross-validation to evaluate the robustness and generalization of the model across different subsets of data.

In addition to supervised classification, these embeddings can also be used in conjunction with a human operator where the embeddings facilitate the initial sorting and grouping of documents, and users refine and verify the results. This strategy allows operators to efficiently navigate and manage vast archives, improving the speed and accuracy of document processing.

Specifically, there are two key aspects we may be interested in: first, whether the neighborhood around a given sample contains documents of the same class, which is useful for similarity search tasks. Second, whether we can derive reliable clusters of document types to quickly and accurately assign a label to a cluster of documents.

The first objective can be achieved by selecting a distance metric and evaluating how often each instance is the same class as its nearest neighbors. To evaluate this across a dataset, we use K-Nearest Neighbors (KNN) classification accuracy using cosine distance. However, because distance metrics can become less discriminative in high-dimensional spaces~\cite{beyerWhenNearestNeighbor1999}, we employ Uniform Manifold Approximation and Projection (UMAP) to reduce dimensionality before applying KNN. We experiment with reducing dimensions to 10, 20, and 30, and report the average results to account for variance. We set the number of neighbors to 10, which is 1 fewer than the number of instances in the smallest class, to ensure that each class had sufficient examples to populate the neighborhood completely if correctly clustered. Additionally, we used Euclidean distance with a minimum distance of 0.1 to control the tightness of point packing in the embedding space.

The second objective is to evaluate the formation of reliable clusters of document types. For this, we choose K-means clustering due to its simplicity and effectiveness in partitioning data into distinct groups based on similarity. Similar to KNN, K-means can be less effective in high-dimensional spaces~\cite{steinbachChallengesClusteringHigh2004}, so we use the same UMAP processing as with KNN before clustering. K-means is applied with Euclidean distance, normalized data, and three trials to ensure stability. The number of clusters is set to reflect the number of form types identified in our datasets.

\subsubsection{Evaluation Metrics}
Because we have the ground truth labels, we can assess the clustering performance using the following metrics:
\begin{itemize}
    \item \textbf{V-measure:} A harmonic mean of completeness and homogeneity, adjusted for chance, which assesses the quality of clustering. This measure is somewhat akin to the \(F_{1}\) score used in classification tasks, but balances two aspects of cluster quality in a single metric.
    \item \textbf{Adjusted Rand Index (ARI):} Measures the similarity between two data clusterings, adjusting for chance grouping.
\end{itemize}

\subsection{Datasets}
Because no existing datasets adequately address this task, we have created two novel datasets specifically designed to test and refine our unsupervised document classification methodology. These datasets are:

\begin{itemize}
    \item \textbf{U.S. 1950 Census Dataset:} Completed census forms from the U.S. 1950 census. It comprises 441 labeled images distributed across 5 form types, with each type having between 48 and 112 examples. For self-supervised MAE training, we train on 9,191 U.S. 1950 Census forms.
    \item \textbf{19th-Century French Census Dataset:} Completed census forms from 19th century France. Contains 591 labeled images distributed across 14 form types, with each type represented by 11 to 61 examples. For self-supervised MAE training, we train on 2,600 instances of similar form types.
\end{itemize}

Determining a form type can be somewhat subjective. For the French Census, a different form type is constituted if any of the following elements differ: column header names, column header name orientations, column width, number of columns, or number of rows. Other variations, such as font style and size, line thickness, or the position of the preprinted form relative to the page, are not considered.

\begin{figure}[h]
  \centering
  {\includegraphics[width=1.0\textwidth]{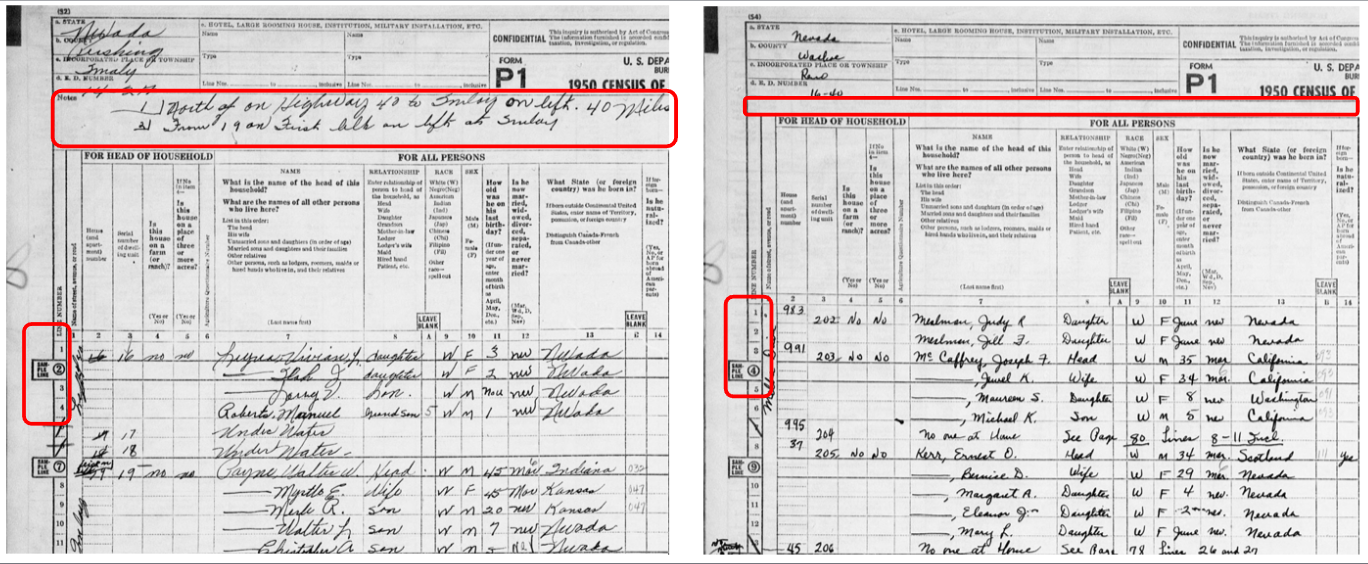}}
  \caption{U.S. 1950 Census forms contain identical content with varying layouts, challenging language-centric models like CLIP to detect subtle differences.
}
  \label{fig:1950s_form_type}

\end{figure}

For the U.S. 1950 Census forms, we consider the variations illustrated in Figure~\ref{fig:1950s_form_type}, which include differences in header row height and numbering outside the first and last columns. The U.S. 1950 Census dataset, characterized by these subtle differences in layout but almost identical content, provides a particularly challenging dataset for representational learning.

\subsection{Model Selection}
We evaluate the representational performance of pretrained models such as DiT, CLIP, and ResNet.

DiT is specifically designed for document image analysis, which suggests it may be well-suited for capturing the structural and semantic features of our datasets. DiT uses average pooling across all encoded patch tokens to compute a 768-dimensional embedding from 224px images. CLIP, which integrates visual and textual data, aims to create robust embeddings by aligning visual features with corresponding text, indicating its potential suitability for document representation. ResNet, pretrained on ImageNet, provides a strong baseline due to its proven effectiveness in various computer vision tasks, offering a reliable comparison point for our document classification methodology.

A pretrained representational model is advantageous from a processing standpoint, as it does not require retraining for each dataset. However, given the diversity of potential data and the likelihood that some data may fall outside the distribution of the pretrained model, these pretrained approaches cannot be relied upon in every case. In such scenarios, self-supervised approaches can offer superior accuracy and adaptability.

Consequently, in addition to evaluating pre-trained models, we also train the self-supervised MAE model on both datasets. The resolution of 448px was chosen over the customary 224px to preserve the distinguishing characteristics of these forms, which are otherwise lost with significant downsampling. Unlike DiT, MAE uses a 768-dimensional class token for downstream tasks.

\subsection{Semantic Segmentation}
\begin{figure}[H]
  \centering
  \includegraphics[trim=0 600 0 0, clip, width=.47\textwidth]{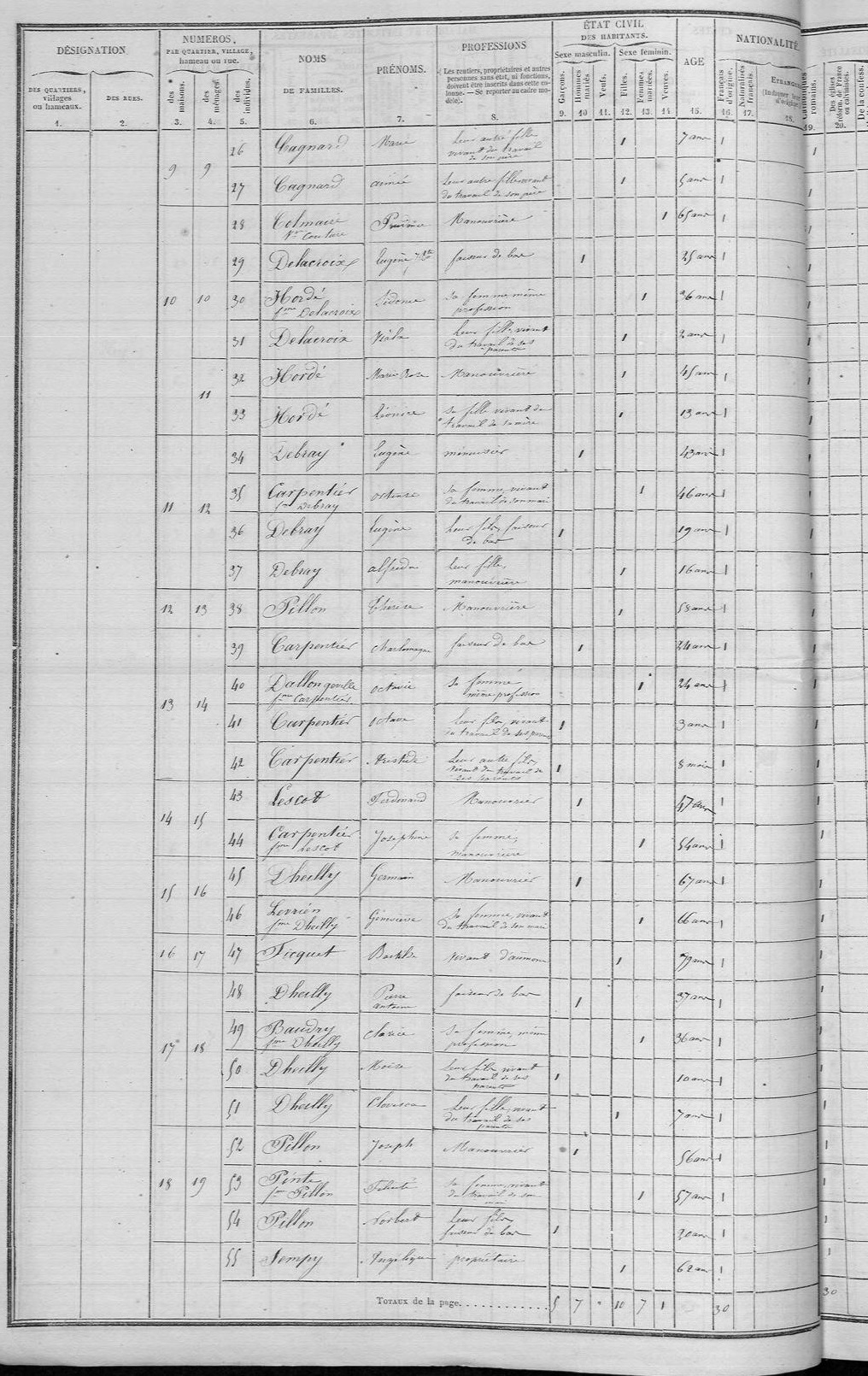}
  \hspace{3mm}
  \includegraphics[trim=0 600 0 0, clip, width=0.47\textwidth]{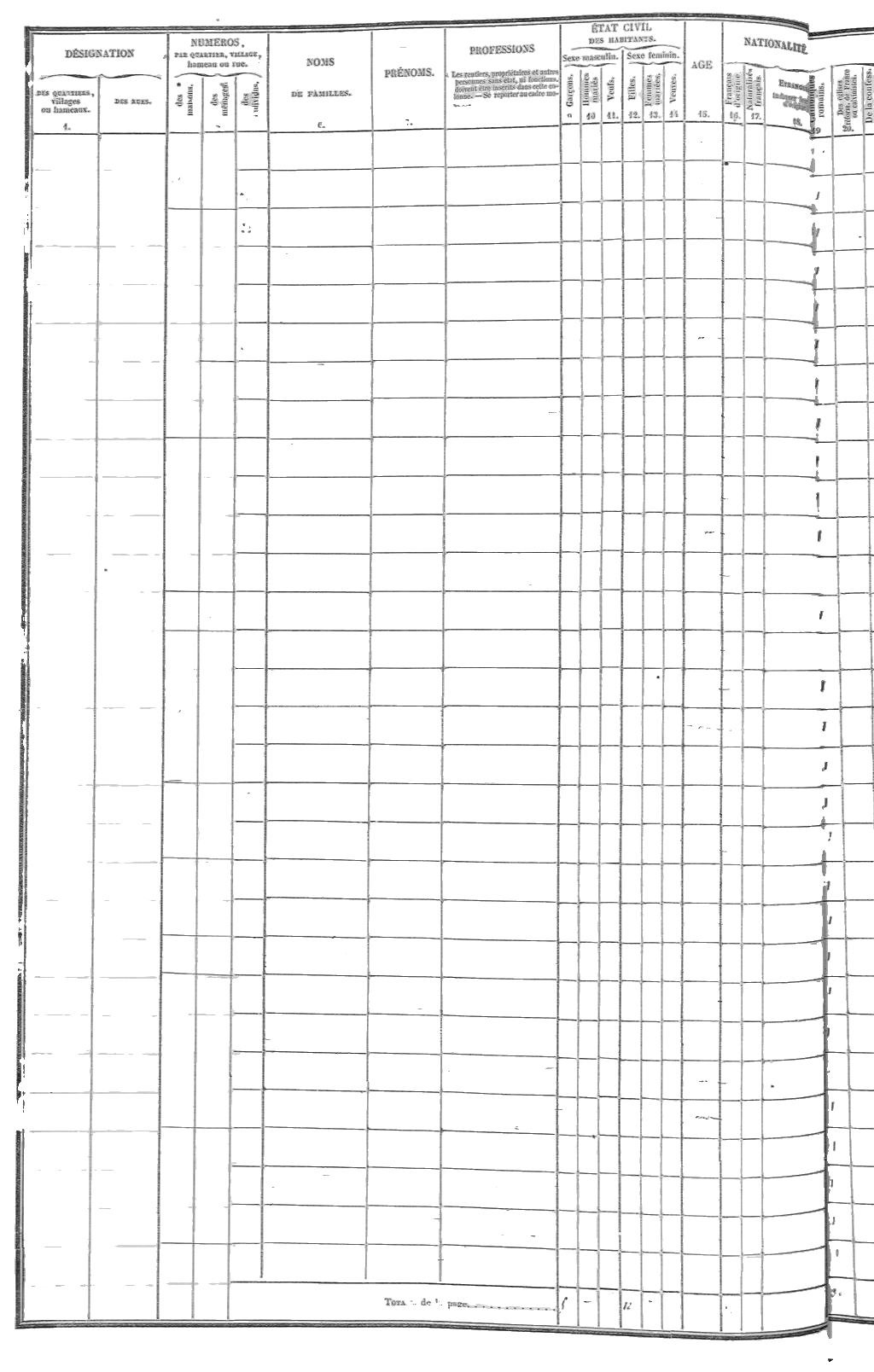}
  \caption{French Census form (left) and its masked counterpart (right) from a model trained on the DELINE8K dataset.}
  \label{fig:french_segmented_before_and_after}
\end{figure}

Given our primary focus on the preprinted elements of the documents, which are crucial for identifying document types, we utilize semantic segmentation to effectively isolate these elements from other content such as handwriting. This segmentation is facilitated by employing a U-Net architecture with a ResNet-50 encoder, trained on the DELINE8K dataset. Figure~\ref{fig:french_segmented_before_and_after} shows a French Census image sample and the corresponding image with the semantic segmentation mask applied.

We specifically apply the predicted masks to preserve form elements and text, masking out all other content including the background and handwriting. We conduct two experiments with these preprocessed images:
\begin{itemize}
\item passing the preprocessed images into pretrained models directly, and
\item training Masked Autoencoders (MAE) on these preprocessed images.
\end{itemize}

These experiments are designed to test the robustness and effectiveness of document classification models when the input data is refined to emphasize structural features over handwritten noise. This approach ensures that our models are tuned to focus on the attributes most relevant to document type classification, thereby enhancing their accuracy and applicability in practical scenarios.

\section{Results and Discussion}
We report the performance of pretrained models, using raw images (No Seg), images with the segmentation masks applied (Seg), and the difference in performance attributable to using the segmented images ($\Delta$ Seg). On the French Census dataset, Table~\ref{tab:french_pretrained} shows that in almost every case the segmented images improve performance, often significantly. The exception to this is CLIP-ViT-L/14-336, which has minor changes in performance in both directions. 

\begin{table}
\caption{Performance metrics on French Census dataset for pretrained models}
\label{tab:french_pretrained}
\centering
\scriptsize
\resizebox{\textwidth}{!}{%
\begin{tabular}{@{}lcccccccccccc@{}}
\toprule
 & \multicolumn{6}{c}{K-Means} & \multicolumn{3}{c}{KNN} & \multicolumn{3}{c}{Linear Probe} \\
\cmidrule(r){2-7}\cmidrule(lr){8-10}\cmidrule(l){11-13}
 & \multicolumn{3}{c}{ARI} & \multicolumn{3}{c}{V-Measure} & \multicolumn{3}{c}{Accuracy} & \multicolumn{3}{c}{Accuracy} \\
Model & No Seg & Seg & $\Delta$ Seg & No Seg & Seg & $\Delta$ Seg & No Seg & Seg & $\Delta$ Seg & No Seg & Seg & $\Delta$ Seg \\
\midrule
CLIP-ViT-B/32 & 0.694 & 0.704 & +0.011 & 0.757 & 0.806 & +0.048 & 0.860 & 0.903 & +0.043 & 0.959 & 0.973 & +0.014 \\
CLIP-ViT-L/14-336 & \textbf{0.833} & 0.809 & \textminus0.024 & \textbf{0.875} & 0.868 & \textminus0.008 & 0.928 & 0.938 & +0.010 & 0.964 & 0.976 & +0.012 \\
DiT-Base & 0.397 & 0.826 & +0.429 & 0.564 & 0.861 & +0.297 & 0.755 & 0.927 & +0.172 & 0.958 & 0.968 & +0.010 \\
DiT-Large & 0.444 & 0.794 & +0.350 & 0.602 & 0.861 & +0.259 & 0.773 & 0.929 & +0.156 & 0.963 & 0.980 & +0.017 \\
ResNet18 & 0.719 & 0.794 & +0.075 & 0.816 & 0.868 & +0.052 & 0.917 & 0.949 & +0.032 & \textbf{0.981} & 0.978 & \textminus0.003 \\
ResNet50 & 0.772 & \textbf{0.864} & +0.092 & 0.853 & \textbf{0.906} & +0.052 & \textbf{0.948} & \textbf{0.955} & +0.008 & 0.980 & \textbf{0.981} & +0.002 \\

\bottomrule
\end{tabular}
}
\end{table}

Table~\ref{tab:1950_pretrained} shows that performance on the US 1950 Census dataset is generally much lower, suggesting that it is a more difficult task. However, the benefit of using the segmented images is even greater, and every model benefits from using the segmented images in every metric, in some cases considerably, as with ResNet50.

\begin{table}[H]
\caption{Performance metrics on U.S. 1950 Census dataset for pretrained models}
\label{tab:1950_pretrained}
\centering
\scriptsize
\resizebox{\textwidth}{!}{%
\begin{tabular}{@{}lcccccccccccc@{}}
\toprule
& \multicolumn{6}{c}{K-Means} & \multicolumn{3}{c}{KNN} & \multicolumn{3}{c}{Linear Probe} \\
\cmidrule(r){2-7}\cmidrule(lr){8-10}\cmidrule(l){11-13}
 & \multicolumn{3}{c}{ARI} & \multicolumn{3}{c}{V-Measure} & \multicolumn{3}{c}{Accuracy} & \multicolumn{3}{c}{Accuracy} \\
Model & No Seg & Seg & $\Delta$ Seg & No Seg & Seg & $\Delta$ Seg & No Seg & Seg & $\Delta$ Seg & No Seg & Seg & $\Delta$ Seg \\
\midrule
CLIP-ViT-B/32 & 0.011 & 0.029 & +0.018 & \textbf{0.038} & 0.050 & +0.011 & 0.410 & 0.515 & +0.104 & 0.862 & 0.875 & +0.014 \\
CLIP-ViT-L/14-336 & 0.000 & 0.099 & +0.099 & 0.019 & 0.126 & +0.107 & 0.447 & 0.605 & +0.158 & 0.846 & 0.907 & +0.061 \\
DiT-Base & 0.005 & 0.017 & +0.013 & 0.025 & 0.045 & +0.020 & 0.373 & 0.664 & +0.290 & 0.885 & 0.938 & +0.052 \\
DiT-Large & 0.003 & 0.024 & +0.021 & 0.017 & 0.057 & +0.040 & 0.382 & 0.651 & +0.269 & 0.850 & \textbf{0.943} & +0.093 \\
ResNet18 & \textbf{0.015} & 0.166 & +0.151 & 0.030 & 0.241 & +0.210 & 0.484 & 0.688 & +0.204 & 0.895 & 0.924 & +0.030 \\
ResNet50 & 0.000 & \textbf{0.275} & +0.275 & 0.023 & \textbf{0.372} & +0.350 & \textbf{0.555} & \textbf{0.728} & +0.173 & \textbf{0.900} & 0.930 & +0.030 \\
\bottomrule
\end{tabular}
}
\end{table}

It is important to note that the models have different embedding dimension sizes for the linear probe, so linear probe accuracy may not be appropriate for comparison across models. The embedding dimensions are 2048 for ResNet50, 512 for ResNet18, 768 for MAE and DiT, and 512 for CLIP. This, however, does not detract from the $\Delta$ we observe when using the segmented images. Notably, because we have reduced the number of dimensions using UMAP before computing the other metrics, the comparison across models on K-means and KNN metrics is still valid.

\subsection{Self-supervision with MAEs}

While MAE encoders can be pretrained on other datasets, as in the case of DiT, achieving the best performance often requires training on the target dataset. Although this necessitates training a new model for each dataset, it can yield improved performance over pretrained baselines.

For this analysis, models trained on segmented images were evaluated on segmented images, and models trained on raw images were evaluated on raw images, ensuring consistent performance assessment. However, the impact of cross-evaluation—training on one modality and evaluating on the other—remains an area for future work, as it can sometimes improve performance. \begin{samepage}
When training MAE, we evaluate several ablations, including:
\begin{itemize}
    \item training on images translated by up to 10\%,
    \item training on segmented images,
    \item training on segmented images and translating by up to 10\%.
\end{itemize}
\end{samepage}

Table~\ref{tab:french_census_mae} shows the results on the French Census dataset. The findings indicate that MAE models trained on each dataset perform better than the pretrained ones we evaluated. This advantage is likely due, at least in part, to the increased resolution, a benefit that will persist until pretrained transformer models adopt higher-resolution inputs. All models tested show strong KNN and linear probe accuracy, though clustering seems to improve when using segmented images and translation as an augmentation.


\begin{table}
\caption{Performance metrics on French Census dataset using ViT-MAE-448}
\label{tab:french_census_mae}
\centering
\scriptsize
\resizebox{\textwidth}{!}{%
\begin{tabular}{@{}lccccccc@{}}
\toprule
Model & \multicolumn{2}{c}{K-Means} & \multicolumn{1}{c}{KNN} & \multicolumn{1}{c}{Linear Probe} \\
\cmidrule(r){2-3}\cmidrule(lr){4-4}\cmidrule(l){5-5}
 & ARI & V-Measure & Accuracy & Accuracy \\
\midrule
Base model & 0.846 & 0.920 &  \textbf{0.992} & \textbf{0.998} \\
Translation &  0.912 & 0.952 &  \textbf{0.992} &  0.993 \\
Segmentation & 0.920 & 0.953 & 0.980 & 0.990 \\
Segmentation + Translation & \textbf{0.939} & \textbf{0.963} & 0.989 & 0.997 \\
\bottomrule
\end{tabular}
}
\end{table}

\begin{table}[h]
\caption{Performance metrics on U.S. 1950 Census dataset using ViT-MAE-448}
\label{tab:us_census_mae}
\centering
\scriptsize
\resizebox{\textwidth}{!}{%
\begin{tabular}{@{}lccccccc@{}}
\toprule
Model & \multicolumn{2}{c}{K-Means} & \multicolumn{1}{c}{KNN} & \multicolumn{1}{c}{Linear Probe} \\
\cmidrule(r){2-3}\cmidrule(lr){4-4}\cmidrule(l){5-5}
 & ARI & V-Measure & Accuracy & Accuracy \\
\midrule
Base model & 0.343 & 0.471 & 0.874 & 0.973 \\
Translation & 0.442 & 0.528 & 0.855 & 0.964 \\
Segmentation & 0.610 & 0.670 & 0.890 & 0.968 \\
Segmentation + Translation &  \textbf{0.768} &  \textbf{0.837} &  \textbf{0.980} &  \textbf{0.984} \\
\bottomrule
\end{tabular}
}
\end{table}

Table~\ref{tab:us_census_mae} demonstrates that segmentation significantly enhances performance on the U.S. 1950 Census dataset. Although the integration of segmentation with MAE does not uniformly improve performance across all scenarios and may occasionally result in slight detriments, the marked gains observed in this particular dataset suggest that segmentation could serve as a valuable baseline and guide future evaluations and adaptations.

\begin{figure}[h]
  \centering
  \begin{minipage}{0.32\textwidth}
    \includegraphics[width=\textwidth]{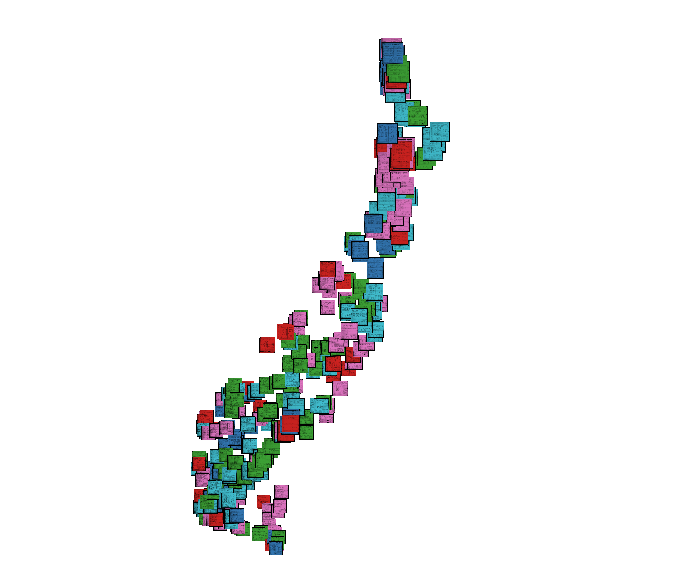}
  \end{minipage}
  \vrule width 0.5pt
  \begin{minipage}{0.32\textwidth}
    \includegraphics[width=\textwidth]{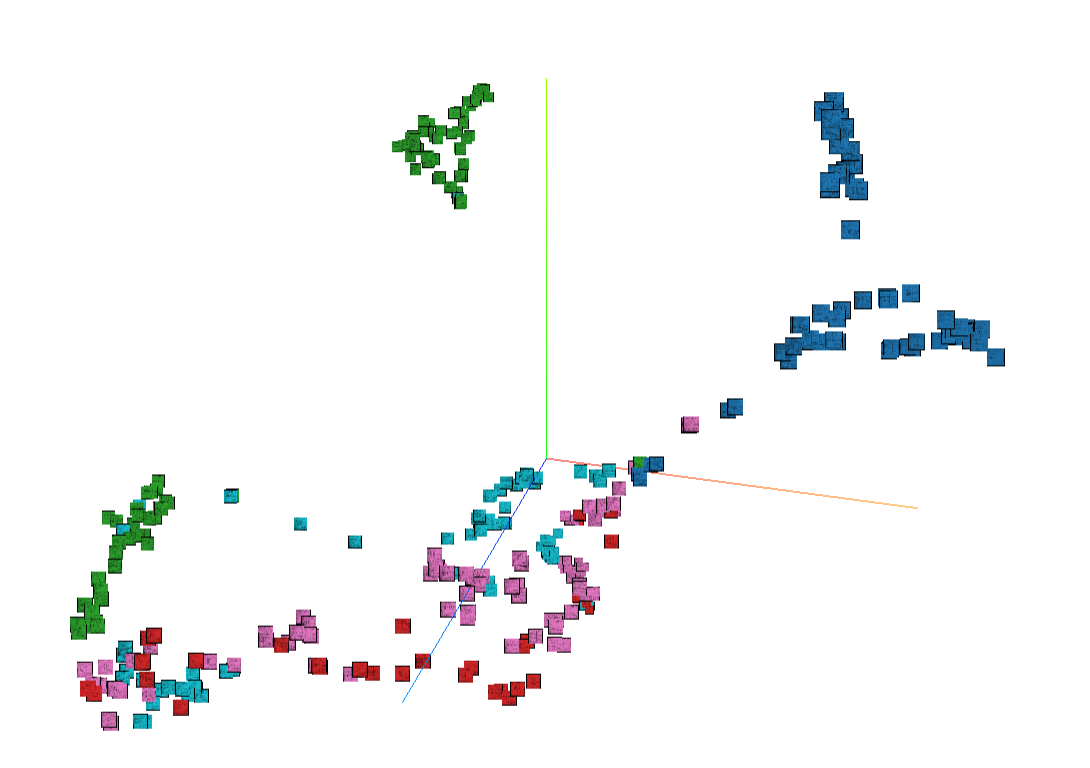}
  \end{minipage}
  \vrule width 0.5pt
  \begin{minipage}{0.32\textwidth}
    \includegraphics[width=\textwidth]{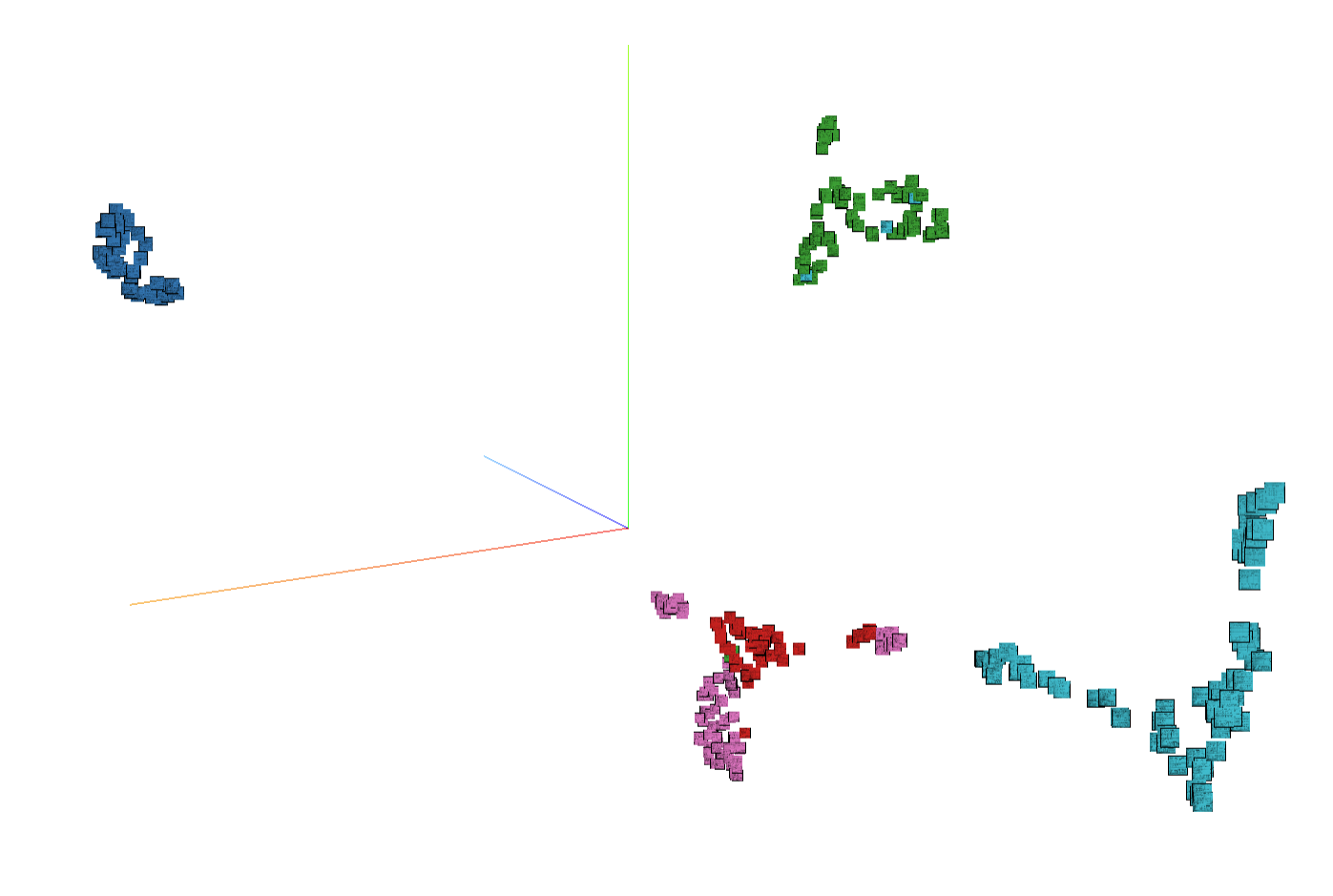}
  \end{minipage}
  \caption{Projection of the U.S. 1950 Census dataset using UMAP based on embeddings from CLIP-ViT-L/14-336 (left), ViT-MAE-448 (middle), and ViT-MAE-448 trained on segmented images (right).}
  \label{fig:projector_view}
\end{figure}

Figure~\ref{fig:projector_view} shows a visualization of the MAE embeddings projected into 3D space. These visualizations confirm the finding that MAE models pretrained on the U.S. 1950 Census dataset produce better representations than pretrained models like CLIP, and furthermore, that these clusters may be further improved by training the model on segmented images.

\section{Future Work}
An area of future work includes investigating other self-supervised methods in conjunction with segmentation masks. Many self-supervised methods, such as SwAV~\cite{caronUnsupervisedLearningVisual2021}, SimCLR~\cite{chenSimpleFrameworkContrastive2020}, Barlow Twins~\cite{zbontarBarlowTwinsSelfSupervised2021}, BYOL~\cite{grillBootstrapYourOwn2020}, and I-JEPA~\cite{assranSelfSupervisedLearningImages2023a}, rely on creating two valid representations of the same data. These methods generally involve creating pairs of augmented views of the same image, encouraging the model to learn invariant features. Given that both the original image and its semantic segmentation are valid representations of the same form type, we propose defining a continuum of valid form representations by interpolating between the original image and its segmented counterpart. This interpolation could help the model learn robust features that are invariant to different representations of the same document, potentially enhancing the model’s ability to generalize across varying document formats and conditions. By leveraging the relationship between the original and segmented images, we aim to improve the effectiveness of self-supervised learning methods in capturing the nuanced structural characteristics of historical documents.

\section{Conclusion}
In this work, we have proposed a representational learning strategy that leverages semantic segmentation and deep learning models to classify historical document images based on their structural characteristics. To the best of our knowledge, this is the first study to evaluate embeddings on fine-grained form types and employ semantic segmentation for form classification. 

We have introduced two novel datasets—the French 19th-century Census and the U.S. 1950 Census records—which serve as challenging benchmarks for fine-grained document classification tasks. Our experimental results demonstrate that the use of semantically segmented images significantly improves the performance of various deep learning models, including ResNets, CLIP, DiT, and MAE. 

These contributions establish a new benchmark in the field of unsupervised fine-grained document classification and open avenues for future research, including the exploration of other self-supervised methods in conjunction with segmentation masks to further enhance document representation.

\section*{Acknowledgment}

The authors thank the Computer Vision Team at Ancestry.com for providing support and the French and U.S. 1950 Census data that contributed to this study. 

\newpage
{
    \small
    \clearpage
    \printbibliography
    \newpage
}
\newpage

\end{document}